\documentclass{article}
\usepackage[final]{neurips_2021}
\usepackage[utf8]{inputenc}
\usepackage[T1]{fontenc}
\usepackage[hidelinks]{hyperref}
\usepackage{url}
\usepackage{booktabs}
\usepackage{amsfonts}
\usepackage{nicefrac}
\usepackage{microtype}
\usepackage[pdftex]{graphicx}
\usepackage{xcolor}
\definecolor{color3}{rgb}{1.0, 0.0, 0.00}
\definecolor{color4}{rgb}{0.0, 0.0, 1.00}
\usepackage{amsmath}
\usepackage{soul}
\usepackage{cleveref}
\usepackage{multirow}
\usepackage{multicol}

\usepackage{subcaption}
\title{\textsc{PreDisM}: Pre-Disaster Modelling With CNN Ensembles for At-Risk Communities}

\author{%
  Vishal Anand\\
  Microsoft Corporation\\
  Redmond, Washington, USA \\
  \texttt{vishal.anand@columbia.edu} \\
  \And
  Yuki Miura \\
  Columbia University\\
  New York, New York, USA \\
  \texttt{yuki.miura@columbia.edu} \\
}

\begin{document}

\maketitle

\begin{abstract}
The machine learning community has recently had increased interest in the climate and disaster damage domain due to a marked increase in occurrences of natural hazards (e.g., hurricanes, forest fires, floods, earthquakes). However, not enough attention has been devoted to mitigating probable destruction from impending natural hazards. We explore this crucial space by predicting building-level damages on a before-the-fact basis that would allow state actors and non-governmental organizations to be best equipped with resource distribution to minimize or preempt losses. We introduce \textsc{PreDisM} that employs an ensemble of ResNets and fully connected layers over decision trees to capture image-level and meta-level information to accurately estimate weakness of man-made structures to disaster-occurrences. Our model performs well and is responsive to tuning across types of disasters and highlights the space of preemptive hazard damage modelling.

\end{abstract}

\section{Introduction}
\label{intro}
\subsection{Natural Hazard Threats and Climate Risk}
Natural hazards 
have been humanity's threats throughout the world over the years, and such threats would worsen as the climate changes \citep{tippett2018extreme}. Specifically, sea-level rise could amplify flooding threats and leave low-elevated coastal regions under sea-level permanently \citep{marsooli2019climate,miura2021high}; warm and dry conditions cause severe and vast wildfires, and such fires could frequently occur as global warming extends fire seasons in the future \citep{halofsky2020changing}; more extreme rainfalls are expected due to warming climate \citep{knutson2015global} and the slow-downed tropical cyclones which are also caused by global warming \citep{kossin2018global}. These threats are some of the many outcomes of changing climate. The regions with little or no hazard occurrences in the past may be devastated by amplified hazards \citep{miura2021optimization}. Such natural hazards affect millions of people and cost billions of dollars \citep{he2018modeling,kreibich2019preface}. Given the limited resources, it is critical to predict and prepare effectively for future hazards in order to mitigate their threats \citep{miura2021method}. Understanding the nature of threats  (i.e., magnitude, frequency) could allow proper preparedness and a quicker recovery. For example, stakeholders and communities could implement a proper protective measure (e.g., barriers) or strategy (e.g., evacuation plan, insurance) against future hazards.


\subsection{Hazard Damage Studies with Machine Learning}
The machine-learning community has been very interested in disaster damage domain in the context of climate change. For example, Crisis-DIAS from \cite{Agarwal_Leekha_Sawhney_Shah_2020} uses multi-modal platforms to consolidate post-disaster damage-levels across information sources, while \cite{Zhang_Huang_Zhang_Wang_2020} discusses human-in-loop entities to accurately estimate post-disaster damages. The work in  \cite{chen2020interpretability} explores convolution neural networks (CNNs) and hyper-parameters to merge pre-disaster and post-disaster images together for the same.

Unlike most previous studies that use machine learning in the space of building damage and climate change \citep{Agarwal_Leekha_Sawhney_Shah_2020, Zhang_Huang_Zhang_Wang_2020, chen2020interpretability}, which focus on after-the-fact damage estimation of a natural hazard occurrence, we introduce \textsc{PreDisM}, which models susceptibility of buildings, landscapes, and habitats for a future calamity. Forecasting the future risks allows residents and governments to take stock of situations with an appropriate contingency fund, facilities, and preparedness. \textsc{PreDisM} doesn't depend on disaster occurrence, rather it predicts damages that could happen from a future hazard with a given intensity. This is highly beneficial to affected communities if the future hazard risk, given the climate risk, could be identified without the massive past data but with the current community's information (e.g., present satellite imagery over the area). For instance, identifying building damage levels prior to a hazard of an arbitrary magnitude hitting the area would provide an insight to the stakeholders and communities. Our models are fine-tuned on disaster-type open-source building datasets \citep{Gupta_2019_CVPR_Workshops}, along with adjoining unifying fully connected layers activated with meta-information to generate building-level hazard-level risks of future damage.

\section{Methodology}
\label{methodology}
\subsection{Data Schema}
\label{data-schema}
We use the xBD dataset from \cite{Gupta_2019_CVPR_Workshops} that serve as the basis of fine-tuning our ensemble model. 
The xBD data is spread across fifteen countries and 45,361 square kilometers that equate to 850,736 individual buildings' data. The 30 GB of data is manually appended by quite a few meta-information (additional details in Appendix). 

Additional information is gleaned for xBD building images that include damage-levels incurred on past occurrences serving the purpose of training our model ensemble predictor. Each damage-level is classified into one of "unclassified", "no damage", "minor damage", "major damage", or "destroyed". During inference using \textsc{PreDisM}, when enough information is not available to make a classification, we would mark them as unclassified. However, we discard such data from our training samples so as to avoid model from accidentally misdiagnosing buildings' risk levels in the absence of enough information. Despite the presence of post-disaster xBD images, we can't use them in our modelling, since our problem requires before-the-fact building-damage modelling on any given satellite imagery.

\subsection{Data treatment}
\label{data-treatment}
\Cref{fig:diagram-a} depicts our data processing flow, while \Cref{fig:diagram-b} describes how granularity is maintained across training and inference. Thirty GB worth of data samples from xBD are split into disaster-types and images within each types are grouped into event-types  (e.g., individual hazard occurrences). 
Each event's image in the dataset comes tagged with building boundaries, that can also be auto-generated for unseen images using \cite{8575504}. Each given image of dimensions (x$\times$y) may contain \textit{n} number of buildings (potentially hundreds), wherein each building-boundary creates separate individual images consisting of just their corresponding  buildings using image masks - so \textit{n} building masks create \textit{n} individual images with (x$\times$y) dimensions of stand-alone buildings to ensure our model does not cross associate damage levels of nearby buildings. This ensures tight coupling of individual buildings with corresponding damage level prediction, as opposed to general damage prediction in a geography. This also ensures our dataset is amplified despite the lack of ample data.

\begin{figure*}[ht]
    \centering
    \vskip -0.1in
    \begin{subfigure}[t]{0.32\textwidth}
        \centering
        \includegraphics[width=\columnwidth]{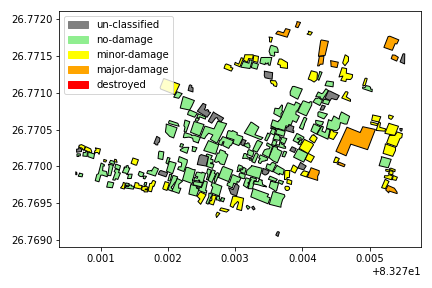}
        \vskip -0.05in
        \caption{Hazard Level 3}
    \end{subfigure}%
    ~ 
    \begin{subfigure}[t]{0.32\textwidth}
        \centering
        \includegraphics[width=\columnwidth]{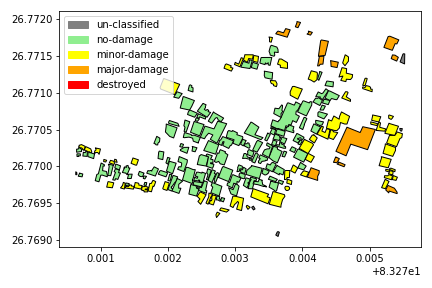}
        \vskip -0.05in
        \caption{Hazard Level 4}
    \end{subfigure}%
    ~ 
    \begin{subfigure}[t]{0.32\textwidth}
        \centering
        \includegraphics[width=\columnwidth]{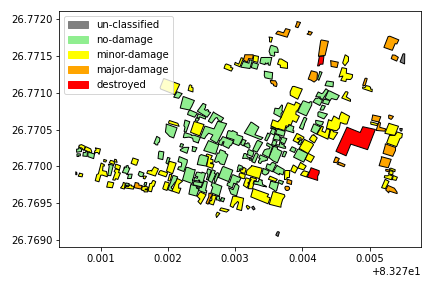}
        \vskip -0.05in
        \caption{Hazard Level 5}
    \end{subfigure}%
    \caption{Qualitative flooding damage prediction ($x=$lat, $y=$lng) without earlier disasters: Nepal}
    \label{fig:damage-level-variation-output}
    \vskip -0.1in
\end{figure*}

`Disaster-types' include earthquake, fire, flood, hurricane, tornado, tsunami, and volcanic eruption and `disaster-levels' vary from 1 to 5, where 5 is the worst case. These levels are defined using impact values in \cite{federal2011risk}, accounting for factors like fatality/injury rate, impaired area, economic loss and supply disruption. The mean of meta-inputs represents the overall hazard level and is appended to our data processing pipeline.
During inference, since hazards haven't occurred, we use varying hazard-levels to create future damage maps (\Cref{fig:damage-level-variation-output}).

\subsection{Modelling}
\label{modelling}
We take the Residual Network (ResNet) \citep{7780459} class of convolutional neural networks that are initially pre-trained on the ImageNet dataset \citep{Krizhevsky2012ImageNetCW} comprised of multi-faceted set of 14 million images. Multiple copies of these ResNet models are then taken and joined together via a decision tree, such that each ResNet model is best responsive to corresponding disaster types that we achieve through fine-tuning. We use multiple ResNet sized models with corresponding loss functions such as cross-entropy, and ordinal-cross-entropy during fine-tuning. Our setup comprised of RTX 2080-Ti (12 GiB) and v3-8 TPU-node (8 TPU cores, 128 GiB). As our experiments evolved across ResNet-18, ResNet-34 and ResNet-50, epoch times increased significantly (e.g., ResNet-34 took 76 minutes per epoch that equates to 25.33 hours for 20 epochs). Thus, we perform experiments with the first two ResNets using Adam optimizer with learning-rate 0.001, gamma of 0.1, learning-rate-decay step-size of 7, and trained over 20 epochs.

\begin{figure*}[ht]
    \centering
    \begin{subfigure}[t]{0.535\textwidth}
        \centering
        \includegraphics[width=\columnwidth]{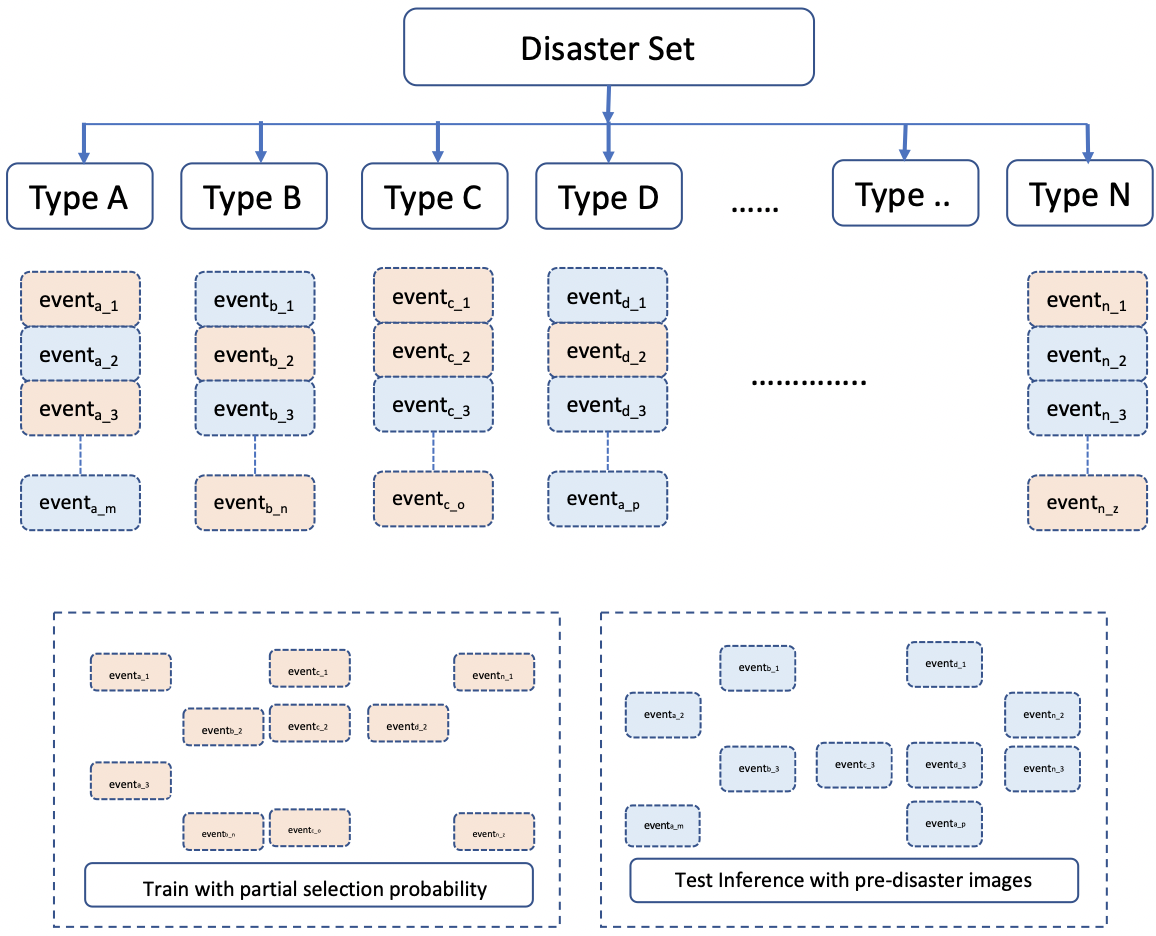}
        \caption{Data Schema Flow}
        \label{fig:diagram-a}
    \end{subfigure}%
    ~ 
    \begin{subfigure}[t]{0.34\textwidth}
        \centering
        \includegraphics[width=\columnwidth]{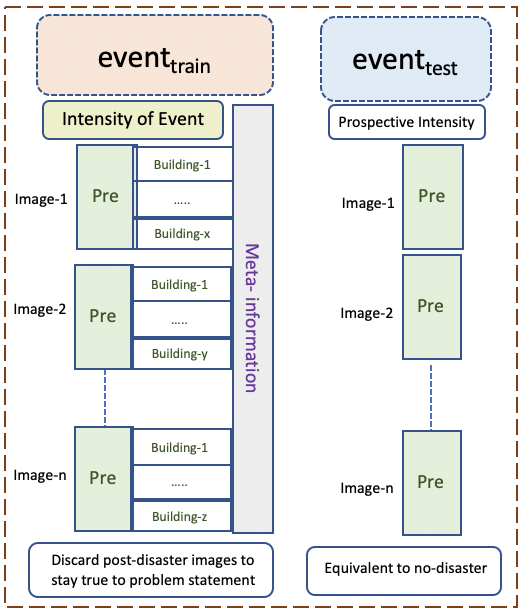}
        \caption{Model data usage}
        \label{fig:diagram-b}
    \end{subfigure}
    
    \begin{subfigure}[t]{\textwidth}
        \centering
        \vskip 0.1in
        \includegraphics[width=0.8\columnwidth]{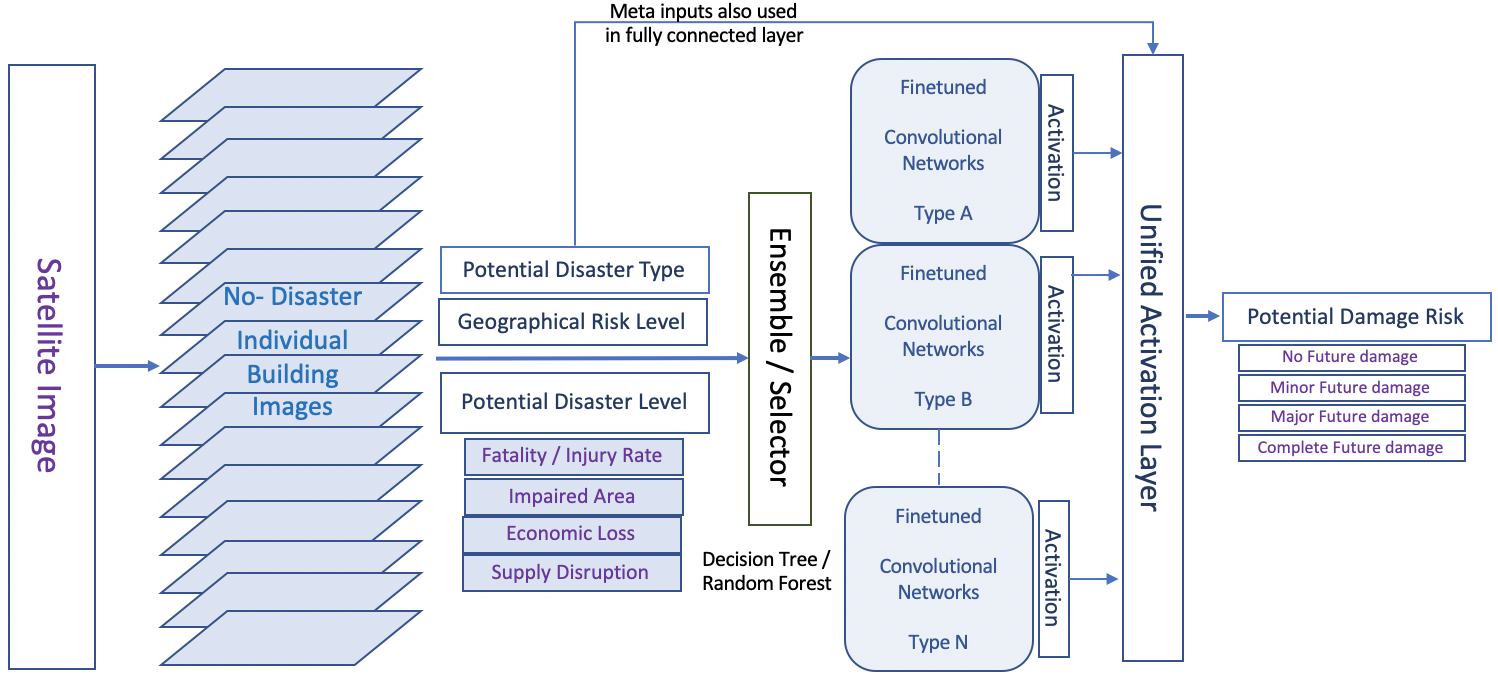}
        \caption{Model inference on non-disaster images (meta-details in Appendix)}
        \label{fig:diagram-c}
    \end{subfigure}
    \caption{Schematic diagram of the data in context of inference.}
    \label{fig:diagram}
\end{figure*}

\Cref{fig:diagram-c} is the schematic representation of our model ensemble. 
Firstly, an input satellite image is parsed into individual building masks that are applied independently to produce multiple images having one-building-per-image. Then, the `suggested' input from climate-scientists comprising of meta-information (such as disaster-type, fatality, land impairment, disruption time) and corresponding future hazard-level decide which fine-tuned ResNet models to select. The ensemble model allows for multiple fine-tuned ResNet models to kick-in (e.g., if two natural hazards usually co-occur, ensemble would assign importance-weights to these two ResNets as learnt from past data). Finally the outputs of ResNet CNN are activated through softmax layers to output damage-levels ranging from 1-5.

We found that rather than adding meta-data to pixel data as mentioned in \cite{chen2020interpretability} which would eventually get diluted; we should process them outside of ResNets. This leads to a marked accuracy increase since meta-information is already in condensed format and in a different vector-space. Thus, we combine them with the ResNet output after passing through fully connected layers to produce the final damage-level forecast. This allows accurate fine-tuning of the entire ensemble, else we would need much larger dataset before a ResNet learns that the appended meta-input is, in fact, not a pixel.

\section{Results and Discussion}
\label{results+discussion}
 \textsc{PreDisM} achieves 79.24\% accuracy (\Cref{tab:result}) and has a marked improvement in future building damage predictions, despite the inherent hardness of the problem since we have no disaster data during inference, and are carrying out preemptive modelling of future hazard of a given intensity.  \Cref{fig:damage-level-variation-output} is the predicted output format for these varying hazard levels. 
 
 Despite the core differences in the problem being investigated by \cite{chen2020interpretability,Agarwal_Leekha_Sawhney_Shah_2020,Zhang_Huang_Zhang_Wang_2020}, we realize \cite{chen2020interpretability} is a reasonable comparison for our experiments since they discuss the aspects on a pre-disaster level. An interesting point to note is the improvement gained over \cite{chen2020interpretability} by our altered data treatment. We use individual building-masks to create corresponding images of same input sizes, doing away with the uneven building size problem. We also use meta-data like disaster-types, hazard-levels with decision tree and fully connected layers as opposed to appending directly into building images, since they are inherently in different data-representation vector-spaces because meta-information are condensed event representations. These are not suitable for CNNs like ResNets, while image-data are on a red-blue-green level which may or may not yield actionable information. Accuracy calculation leverages gold (post-disaster) data only for checking model performance, and never during inferencing/training of input image.

\textsc{PreDisM}'s output in \Cref{fig:damage-level-variation-output} would serve as an excellent guide to officials to prepare for impending disasters for varying levels of hazards over frequent satellite image-datasets. The affected community's damage could significantly vary depending on the ground surface property. 
For example, the presence of wetlands could work as a bumper by slowing down currents and reducing the impact of tsunami and floods considerably. Consequently, it is easy to imagine the corresponding damage would be minor compared to the area without such a bumper. For the future model, it is encouraged to include such geographical features as well. Our results suggest a pre-disposition of buildings and area-conditions to have predictive damage-risk that can be analyzed with manually-controlled potential disaster's intensities to better prepare for unforeseen calamity and avoidable anguish.

\begin{table*}[h]
\centering
\small
\begin{tabular}{lccc}
\toprule
\multirow{1}{*}{\textsc{Model}} &
\multirow{1}{*}{\textsc{Loss-Function}}&
\multicolumn{1}{c}{\textsc{Accuracy}}\\
\midrule
\multirow{1}{*}{\textsc{PreDisM$_{\text{ResNet-18}}$}} 
& Cross-Entropy & 78.38 \% \\
\midrule
\multirow{1}{*}{\textsc{PreDisM$_{\text{ResNet-34}}$}} 
& Cross-Entropy & 79.24 \% \\
\midrule
\multirow{2}{*}{Chen$_{post}$} 
& Cross-Entropy & 59.50 \% \\
& Ordinal Cross-Entropy & 64.20 \% \\
\hline
\end{tabular}
\caption{\label{tab:result} Prediction inference on non-disaster images}
\vskip -0.25in
\end{table*}

\section{Conclusion}
Our work introduces a new paradigm of working towards climate damage preemptive preparedness. \textsc{PreDisM} uses disaster-type specific fine-tuned convolution neural networks of ResNets wrapped into ensembles over fully connected layers and decision trees on non-disaster afflicted satellite images, along with adaptive damage-levels map-outputs sensitive to potential hazard intensity from government officials. This demonstrates marked accuracy gains over other models in similar settings and would serve as a guideline to decrease damage potential on a rolling basis. Stakeholders and administrators can effectively plan mitigations for future disasters using our model. 


{
\small
\bibliographystyle{apalike}
\bibliography{references}

\begin{thebibliography}{}

\bibitem[Agarwal et~al., 2020]{Agarwal_Leekha_Sawhney_Shah_2020}
Agarwal, M., Leekha, M., Sawhney, R., and Shah, R.~R. (2020).
\newblock Crisis-dias: Towards multimodal damage analysis - deployment,
  challenges and assessment.
\newblock {\em Proceedings of the AAAI Conference on Artificial Intelligence},
  34(01):346--353.

\bibitem[Chen, 2020]{chen2020interpretability}
Chen, T.~Y. (2020).
\newblock Interpretability in convolutional neural networks for building damage
  classification in satellite imagery.
\newblock {\em Tackling Climate Change with Machine Learning workshop at
  NeurIPS 2020}.

\bibitem[{Federal Office of Civil Protection and Disaster Assistance},
  2011]{federal2011risk}
{Federal Office of Civil Protection and Disaster Assistance} (2011).
\newblock {\em Method of Risk Analysis for Civil Protection}.
\newblock Federal Office of Civil Protection and Disaster Assistance.

\bibitem[Gupta et~al., 2019]{Gupta_2019_CVPR_Workshops}
Gupta, R., Goodman, B., Patel, N., Hosfelt, R., Sajeev, S., Heim, E., Doshi,
  J., Lucas, K., Choset, H., and Gaston, M. (2019).
\newblock Creating xbd: A dataset for assessing building damage from satellite
  imagery.
\newblock In {\em Proceedings of the IEEE/CVF Conference on Computer Vision and
  Pattern Recognition (CVPR) Workshops}.

\bibitem[Halofsky et~al., 2020]{halofsky2020changing}
Halofsky, J.~E., Peterson, D.~L., and Harvey, B.~J. (2020).
\newblock Changing wildfire, changing forests: the effects of climate change on
  fire regimes and vegetation in the pacific northwest, usa.
\newblock {\em Fire Ecology}, 16(1):4.

\bibitem[He et~al., 2016]{7780459}
He, K., Zhang, X., Ren, S., and Sun, J. (2016).
\newblock Deep residual learning for image recognition.
\newblock In {\em 2016 IEEE Conference on Computer Vision and Pattern
  Recognition (CVPR)}, pages 770--778.

\bibitem[He and Cha, 2018]{he2018modeling}
He, X. and Cha, E.~J. (2018).
\newblock Modeling the damage and recovery of interdependent critical
  infrastructure systems from natural hazards.
\newblock {\em Reliability engineering \& System safety}, 177:162--175.

\bibitem[Knutson et~al., 2015]{knutson2015global}
Knutson, T.~R., Sirutis, J.~J., Zhao, M., Tuleya, R.~E., Bender, M., Vecchi,
  G.~A., Villarini, G., and Chavas, D. (2015).
\newblock Global projections of intense tropical cyclone activity for the late
  twenty-first century from dynamical downscaling of cmip5/rcp4. 5 scenarios.
\newblock {\em Journal of Climate}, 28(18):7203--7224.

\bibitem[Kossin, 2018]{kossin2018global}
Kossin, J.~P. (2018).
\newblock A global slowdown of tropical-cyclone translation speed.
\newblock {\em Nature}, 558(7708):104--107.

\bibitem[Kreibich et~al., 2019]{kreibich2019preface}
Kreibich, H., Thaler, T., Glade, T., and Molinari, D. (2019).
\newblock Preface: Damage of natural hazards: assessment and mitigation.
\newblock {\em Natural Hazards and Earth System Sciences}, 19(3):551--554.

\bibitem[Krizhevsky et~al., 2012]{Krizhevsky2012ImageNetCW}
Krizhevsky, A., Sutskever, I., and Hinton, G.~E. (2012).
\newblock Imagenet classification with deep convolutional neural networks.
\newblock {\em Communications of the ACM}, 60:84 -- 90.

\bibitem[Marsooli et~al., 2019]{marsooli2019climate}
Marsooli, R., Lin, N., Emanuel, K., and Feng, K. (2019).
\newblock Climate change exacerbates hurricane flood hazards along us atlantic
  and gulf coasts in spatially varying patterns.
\newblock {\em Nature communications}, 10(1):1--9.

\bibitem[Miura et~al., 2021a]{miura2021optimization}
Miura, Y., Dinenis, P.~C., Mandli, K.~T., Deodatis, G., and Bienstock, D.
  (2021a).
\newblock Optimization of coastal protections in the presence of climate
  change.
\newblock {\em Frontiers in Climate}, 3:83.

\bibitem[Miura et~al., 2021b]{miura2021high}
Miura, Y., Mandli, K.~T., and Deodatis, G. (2021b).
\newblock High-speed gis-based simulation of storm surge--induced flooding
  accounting for sea level rise.
\newblock {\em Natural Hazards Review}, 22(3):04021018.

\bibitem[Miura et~al., 2021c]{miura2021method}
Miura, Y., Qureshi, H., Ryoo, C., Dinenis, P.~C., Li, J., Mandli, K.~T.,
  Deodatis, G., Bienstock, D., Lazrus, H., and Morss, R. (2021c).
\newblock A methodological framework for determining an optimal coastal
  protection strategy against storm surges and sea level rise.
\newblock {\em Natural Hazards}, 107(2):1821–1843.

\bibitem[Tippett, 2018]{tippett2018extreme}
Tippett, M.~K. (2018).
\newblock Extreme weather and climate.

\bibitem[Zhang et~al., 2020]{Zhang_Huang_Zhang_Wang_2020}
Zhang, D.~Y., Huang, Y., Zhang, Y., and Wang, D. (2020).
\newblock Crowd-assisted disaster scene assessment with human-ai interactive
  attention.
\newblock {\em Proceedings of the AAAI Conference on Artificial Intelligence},
  34(03):2717--2724.

\bibitem[Zhao et~al., 2018]{8575504}
Zhao, K., Kang, J., Jung, J., and Sohn, G. (2018).
\newblock Building extraction from satellite images using mask r-cnn with
  building boundary regularization.
\newblock In {\em 2018 IEEE/CVF Conference on Computer Vision and Pattern
  Recognition Workshops (CVPRW)}, pages 242--2424.

\end{thebibliography}
}

\newpage
\appendix

\section{Appendix}
\label{appendix}
\begin{table}[h]
\small
\caption{Hazard Dataset.}
\label{tab:hazardtype}
\vskip 0.15in
\begin{center}
\begin{sc}
\begin{tabular}{lll}
\toprule
\textbf{Hazard Type} & \textbf{Location/Name} & \textbf{Year} \\
\midrule
Earthquake                  & Mexico City       & 2017 \\
\midrule
\multirow{4}{*}{Wildfire}   & Santa Rosa, CA    & 2017 \\
                            & Pinery            & 2015 \\
                            & Portugal          & 2017 \\
                            & Woolsey, CA       & 2018 \\
\midrule
\multirow{2}{*}{Flood}      & Midwest, US       & 2019 \\
                            & Nepal             & 2017 \\
\midrule
\multirow{4}{*}{Hurricane}  & Florence          & 2018 \\
                            & Harvey            & 2017 \\
                            & Matthew           & 2016 \\
                            & Michael           & 2018 \\
\midrule
\multirow{3}{*}{Tornado}    & Joplin, MO        & 2011 \\
                            & Moore, OK         & 2013 \\
                            & Tuscaloosa, AL    & 2011 \\
\midrule
\multirow{2}{*}{Tsunami}    & Palu, Indonesia   & 2018 \\
                            & Sunda, Indonesia  & 2018 \\
\midrule
\multirow{2}{*}{Volcanic Eruption} & Guatemala  & 2018 \\
                            & Lower Puna        & 2018 \\
\bottomrule
\end{tabular}
\end{sc}
\end{center}
\vskip 0.1in
\end{table}

\begin{table}[h]
\caption{Hazard Levels as a function of attributes}
\vskip 0.15in
\small
    \centering
    \begin{tabular}{lccccc}
    \toprule
    \multirow{2}{*}{\textsc{Attributes}} & \multicolumn{5}{c}{\textsc{Hazard Level}} \\
            & 5 & 4 & 3 & 2 & 1 \\
    \midrule
        Fatality                    & >10000 & >1000 & >100 & >10 & >1 \\
        Injury                      & >100000 & >10000 & >1000 & >100 & >10 \\
        Land Impaired (km$^2$)      & >500 & >100 & >50 & >10 & >1 \\
        Direct Damage (billion USD) & >100 & >10 & >1 & >0.1 & >0.01 \\
        Indirect Damage (billion USD) & >100 & >10 & >1 & >0.1 & >0.01 \\
        Water Disruption (days)     & >30 & >14 & >7 & >3 & >1 \\
        Energy Disruption  (days)   & >30 & >14 & >7 & >3 & >1 \\
    \bottomrule
    \end{tabular}
    \label{tab:hazardlevel}
\end{table}

\end{document}